\documentclass[final]{cvpr}

\usepackage{times}
\usepackage{epsfig}
\usepackage{graphicx}
\usepackage{amsmath}
\usepackage{amssymb}
\usepackage{booktabs}
\usepackage{multirow}
\usepackage{mathtools}
\include{xcolor}
\usepackage{enumitem}

\usepackage[pagebackref=true,breaklinks=true,colorlinks,bookmarks=false]{hyperref}

\def\cvprPaperID{532} 
\def\confYear{CVPR 2021}

\newcommand{\sampler}{\mathrm{S}}
\begin{document}

\title{Flow Guided Transformable Bottleneck Networks for Motion Retargeting}
\author{Jian Ren\quad Menglei Chai \quad Oliver J. Woodford\thanks{Work done while at Snap Inc.}  \quad Kyle Olszewski\quad Sergey Tulyakov \\
  Snap Inc. \\
{
{\tt\small \{jren, mchai, kolszewski, stulyakov\}@snap.com}}}

\maketitle

\begin{abstract}
Human motion retargeting aims to transfer the motion of one person in a ``driving'' video or set of images to another person.
Existing efforts leverage a long training video from each target person to train a subject-specific motion transfer model.
However, the scalability of such methods is limited, as each model can only generate videos for the given target subject, and such training videos are labor-intensive to acquire and process.
Few-shot motion transfer techniques, which only require one or a few images from a target, have recently drawn considerable attention.
Methods addressing this task generally use either 2D or explicit 3D representations to transfer motion, and in doing so, sacrifice either accurate geometric modeling or the flexibility of an end-to-end learned representation.
Inspired by the Transformable Bottleneck Network, which renders novel views and manipulations of \emph{rigid} objects, we propose an approach based on an implicit volumetric representation of the image content, which can then be spatially manipulated using volumetric flow fields.
We address the challenging question of how to aggregate information across different body poses, learning flow fields that allow for combining content from the appropriate regions of input images of highly \emph{non-rigid} human subjects performing complex motions into a single implicit volumetric representation.
This allows us to learn our 3D representation solely from videos of moving people.
Armed with both 3D object understanding and end-to-end learned rendering, this categorically novel representation delivers state-of-the-art image generation quality, as shown by our quantitative and qualitative evaluations.
\end{abstract}

\section{Introduction}
Retargeting human body motion --- transferring motion from a ``driving'' image or video of one subject (the \emph{source}) to another subject (the \emph{target}), using one or more reference images of the target subject in an arbitrary pose --- has received a great deal of attention in recent years, due to numerous practical and entertaining applications in content generation~\cite{wang2019fewshotvid2vid,Chan_2019_ICCV}.
Such applications include transferring sophisticated athletic techniques or dancing performances to untrained celebrities for special effects for cinema and television; creating amusing performances for one's friends or acquaintances for sheer entertainment; and creating plausible motion sequences from photos or videos depicting famous and important political figures (including historical figures who may no longer be alive to perform such actions) for the creation of plausible full-body ``deepfake'' videos.
However, retaining the target subject's identity while rendering them in novel, unseen poses is highly challenging, and the state-of-the-art is still far from plausible.
 
Many approaches to this task learn to render a specific person~\cite{aberman2019deep,chai2020neural,Chan_2019_ICCV,gafni2019vid2game,isola2017image,knoche2020reposing,ren2020human,wang2018high,wang2018video,wei2020gac,yang2020transmomo,zhou2019dance} conditioned on the desired pose.
This requires a large number of training frames of that person, and incurs substantial training time that must be repeated per each new subject.
By contrast, in the few-shot setting, addressed in this work, only a few reference images of the target are available, and video generation from those images should be fast (\ie, requiring no subject-specific training).
To overcome the lack of data for a given subject, many other techniques~\cite{bhatnagar2019multi,lazova2019360,li2019dense,liu2019liquid,ma2020learning,mir2020learning,neverova2018dense,patel2020tailornet} leverage existing human body models~\cite{alp2018densepose,kanazawa2018end,loper2015smpl} to construct an approximate representation of the subject that can then be manipulated and rendered.
While the 3D nature of these representations often leads to improved performance over their purely 2D counterparts~\cite{balakrishnan2018synthesizing,ma2017pose,ma2018disentangled,ren2020deep-cv,si2018multistage}, their explicit nature, which faces the limitations of capturing salient details with standard human body models, also leads to reduced modeling power and therefore fidelity.
Large variations in the clothing (\eg, dresses or jackets that do not conform to the body shape), body type, or hair of the source and target subjects, for example, cannot easily be represented with standard models that only represent the body itself.

In this work we attain more flexible and expressive modeling power by exploiting a representation that allows for 3D modeling and manipulation, and yet is fully implicit, \ie it can be fully learned, even though we use no explicit ground-truth 3D information such as meshes or voxel grids as supervision.
Recently, just such a representation, the Transformable Bottleneck Network (TBN)~\cite{olszewski2019transformable}, has been shown to produce excellent results on novel view synthesis of rigid objects.
In that work, image content is encoded into an implicit volumetric representation (the ``bottleneck''), in which each of the encoded features in this volume correspond to the local structure and appearance of the corresponding spatial region in the volume depicted in the image.
However, while it requires no 3D supervision, it is trained using multi-view datasets of \emph{rigid} objects depicted from multiple viewpoints to produce implicit volumes that can then be rigidly transformed to produce novel views of the depicted content corresponding to changes in viewpoint.

We build upon this approach to address the challenge of performing motion retargeting for \emph{non-rigid} humans (for which multiple images of a given subject may be available, but in dramatically different poses).
In doing so, we address several challenges: how to aggregate volumetric features from images with changes in camera and body pose, and how to learn this aggregation from videos without explicit 3D or camera pose supervision.
With such an implicit representation, to synthesize a novel pose, we achieve \emph{non-rigid} implicit volume aggregation and manipulation by learning a 3D flow to resample the 3D body model from input images captured with the subject performing various poses or under different viewpoints.
To allow for expressing large-scale motion while retaining fine-grained details in the synthesized images, we propose a multi-resolution scheme in which image content is encoded, transformed and aggregated into bottlenecks of different resolution scales.

As we focus on transferring motion between human subjects, our network pipeline is designed and  trained specifically to extract and manipulate the foreground of the encoded images, with a separate network for extracting and compositing the background with the synthesized result.
Our training scheme employs techniques and loss functions precisely designed for the challenging task of producing plausible motion retargeting without 3D supervision or the use of explicit 3D models, \eg making use of specialized training techniques to teach the network to synthesize plausible results when no ground-truth images corresponding to the applied spatial manipulation is available.
We thus avoid the limitations of explicit body representations~\cite{li2019dense,liu2019liquid}, which may lead to unrealistic results due to the limited reconstruction accuracy and mesh precision.
Furthermore, it allows for learning directly from real 2D images and videos without requiring the tedious and cumbersome collection of copious high-fidelity 3D data~\cite{lazova2019360,neverova2018dense,grigorev2019coordinate}.

In our experiments, we demonstrate that our approach qualitatively and quantitatively outperforms state-of-the-art approaches to human motion transfer, despite the few images used for inference, and even allows for plausible motion transfer when using only a single image of the target.


In summary, our key contributions are:
\begin{itemize}[leftmargin=1.5em]
  \setlength\itemsep{-0.25em}
  \item A novel set of neural network architectures for performing implicit volumetric human motion retargeting, which exploits the power of 3D human motion modeling while avoiding the limitations of standard 3D human body modeling techniques.
  \item A framework to train these networks to attain high-fidelity human motion transfer using only a few example images of the target subject performing various poses, \emph{without} requiring target-specific training.
  \item Evaluations demonstrating our few-shot approach outperforms state-of-the-art alternatives both quantitatively and qualitatively, even those requiring training models for each new subject with substantial training data.
\end{itemize}

\section{Related Work}

\noindent\textbf{Video-to-Video Generation.}
Existing works on video-to-video generation can synthesize high-quality human motion videos using conditional image generation~\cite{chai2020neural,isola2017image,wang2018high}. Chan \etal~\cite{Chan_2019_ICCV} apply pre-computed human poses from driving videos as input for novel view and pose generation of a target person. Along this line, several works improve the  synthesis quality through additional input signals~\cite{gafni2019vid2game}, pose augmentation and pre-processing~\cite{ren2020human,wei2020gac,yang2020transmomo,zhou2019dance}, and temporal coherence~\cite{aberman2019deep,wang2018video}. However, a long recorded video and person-specific training are required for each target person, limiting the scalability of such methods. 
Our work targets a few-shot scenario, with just a handful of source images available, as discussed below.

\begin{figure*}[]
\begin{center}
\includegraphics[width=1\linewidth]{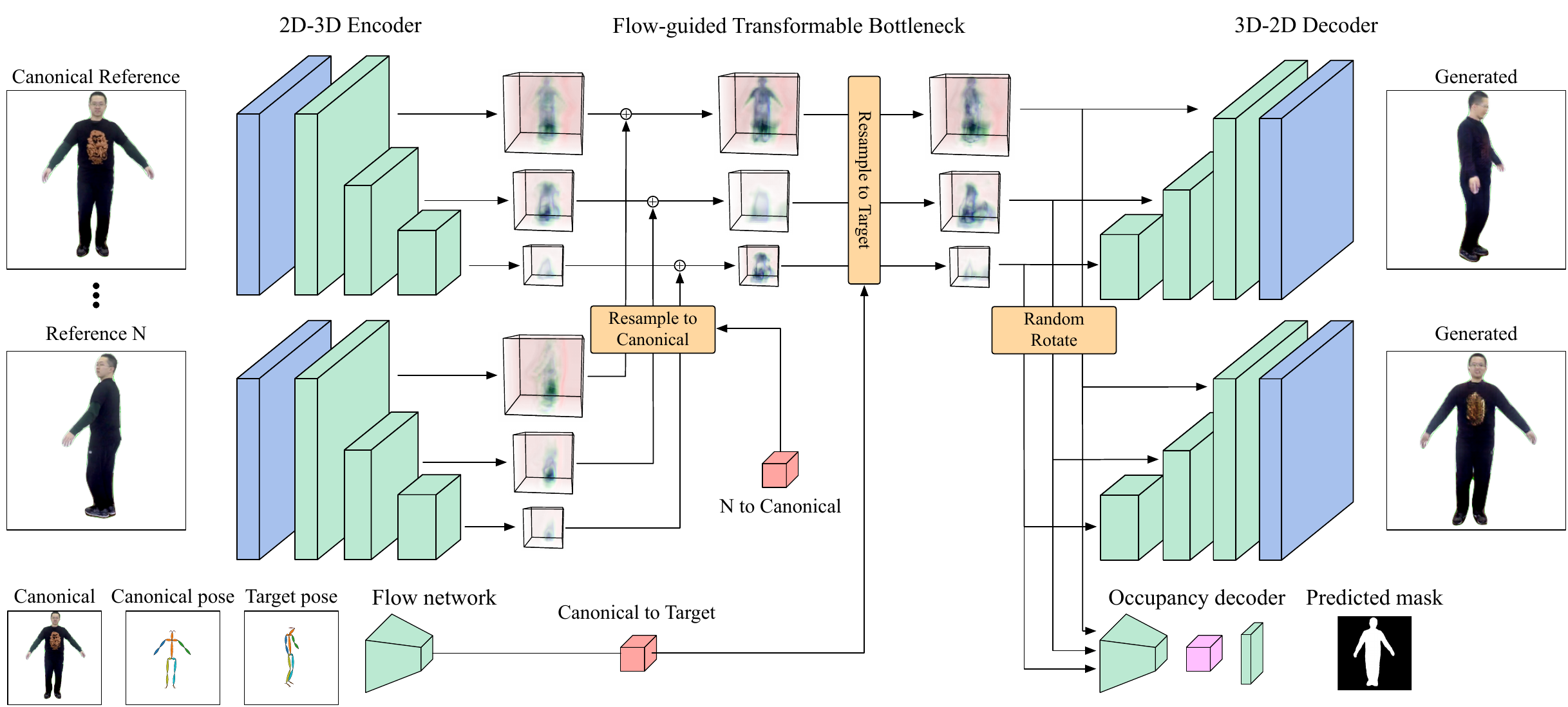}
\caption{Overall pipeline. Given $N$ reference images (left), we extract implicit volumetric representations at multiple scales using our 2D-3D encoder. Our flow network (bottom left) computes the 3D warping field used to warp these representations from the canonical pose (the first source image, to which the other source image feature volumes are aligned) to the target pose. Our 3D-2D decoder then synthesizes the subject in the target pose (top right), while transformations such as rotations can be applied to synthesize novel views of the subject in this pose (middle right). Our occupancy decoder (bottom right) improves the bottleneck's spatial disentanglement and thus allows for better motion retargeting. It computes an occupancy volume used to indicate which regions of the volume are occupied by the subject, which is then decoded into a 2D foreground segmentation mask.}\label{fig:pipline}
\end{center}
\vspace{-2em}
\end{figure*}

\noindent\textbf{Few-Shot Motion Retargeting.} To address scalability, others train generic models that can use just a few images of the target subject in arbitrary poses at test time to synthesize novel images with given poses. While high-quality results have been achieved for face animation~\cite{gu2019flnet,ha2019marionette,qian2019make,zakharov2020fast,zakharov2019few}, animating bodies remains challenging.
Some methods adapt the video-to-video approach to a few-shot setting, using an extra network to generate identity-specific weights for image generation network~\cite{wang2019fewshotvid2vid}, or adapting a pre-trained network to new subjects~\cite{lee2019metapix}.
Another class of methods train networks conditioned on an image of the source identity, and an explicit representation of target pose~\cite{ma2017pose,ma2018disentangled,esser2018variational}.
More recent works exploit multi-stage networks to improve quality~\cite{si2018multistage,ilyes2018pose,dong2018soft}, in particular using 2D spatial transformers~\cite{balakrishnan2018synthesizing,ren2020deep,ren2020deep-cv} or deformation~\cite{siarohin2018deformable} to warp the source image into the target pose, or synthesizing images using attention~\cite{tang2020xinggan,zhu2019progressive}.
However, such purely 2D methods struggle to capture the complex motions generated by 3D shapes and transformations.
Several works use explicit 3D representations, exploiting off-the-shelf human pose~\cite{alp2018densepose} and shape~\cite{kanazawa2018end} inference networks and body meshes~\cite{loper2015smpl}. DensePose~\cite{alp2018densepose} is used to unwrap appearance from the source image(s) into a canonical texture map, which is then inpainted and re-rendered in the target pose~\cite{lazova2019360,neverova2018dense,grigorev2019coordinate}. 
Other works use 3D meshes to compute 2D flow fields to warp features from source to target images~\cite{li2019dense,liu2019liquid}, with shallow encoders and decoders at each end to reduce warping artifacts. Finally, several works use body models and standard rendering techniques for clothing transfer~\cite{bhatnagar2019multi,ma2020learning,mir2020learning,patel2020tailornet}. However, such explicit representations generally produce synthetic-looking results.
We use an \emph{implicit} 3D representation on this task for the first time, unlocking several benefits: the motion representation is more flexible, no 3D supervision or prior is needed, the image decoder can refine the output easily, and multiple reference images can be used to improve synthesis quality.


\noindent\textbf{Unsupervised Motion Retargeting.}
Unsupervised methods learn retargeting purely from videos~\cite{Siarohin_2019_NeurIPS,lathuiliere2020motion,kim2019unsupervised,song2019unsupervised,pumarola2018unsupervised}, forgoing motion supervision and therefore also class-specific motion representations. Siarohin \etal learn unsupervised keypoints in order to warp object parts into novel poses~\cite{siarohin2019animating,Siarohin_2019_NeurIPS}. Lorenz \etal~\cite{lorenz2019unsupervised} show that body parts can also be represented by unsupervised learning, which also helps to disentangle body pose and shape~\cite{esser2019unsupervised}.
Nevertheless, without the benefit of explicit pose information from a human detection model, these methods struggle to generate good results for challenging driving poses. We therefore use a high quality, off-the-shelf, pose detection model~\cite{cao2018openpose} to facilitate the generation process.

\noindent\textbf{Implicit and Volumetric 3D Representations.}
The flexibility of implicit 3D content representations~\cite{10.1145/1015706.1015816,10.1145/383259.383266,survey_implicit_2007},
which obviate the use of explicit surfaces, \eg discrete triangle meshes, make them amenable to recent learning-based approaches that extract the information needed to render or reconstruct scenes directly from images.
In~\cite{sitzmann2019deepvoxels}, an object-specific network is trained from many calibrated images to extract deep features embedded in a 3D grid that are used to synthesize novel views of the object.
\cite{mildenhall2020nerf} perform novel view synthesis on complex scenes by aggregating samples from a trained network which, given a point in space and a view direction, provides an opacity and color value indicating the radiance towards the viewer from this point. %
This approach was later extended to handle unconstrained, large-scale scenes~\cite{martinbrualla2020nerf}, or to use multiple radiance field functions, stored in sparse voxel fields to better capture detailed scenes with large empty regions~\cite{liu2020neural}.
However, these approaches share the limitation that the networks are trained on many images with known camera poses for a specific scene, and thus cannot be used to perform 3D reconstruction or novel view synthesis from new images without re-training.
Furthermore, many images from different viewpoints are required to train these networks to learn to sufficiently render points between these images.
Other recent efforts use 3D feature grids~\cite{kar2017lsm} or pixel-wise implicit functions~\cite{saito2019pifu,saito2020pifuhd,li2020monocular} to infer dynamic shapes from one or more images, but these require synthetic ground-truth 3D geometry for supervision, which limits their applicability to unconstrained real-world conditions.
In~\cite{olszewski2019transformable}, an encoder-decoder framework is used to extract a volumetric implicit representation of image content that is spatially disentangled in a manner that allows for novel view synthesis, 3D reconstruction and spatial manipulation of the encoded image content.
However, while it allows for performing non-rigid manipulations of the image content after training, it must be trained using a multi-view dataset of \emph{rigid} objects, making it unsuitable for human motion retargeting.
In our work we develop an enhanced implicit volumetric representation and multi-view fusion techniques to address these concerns.

\section{Method}

At the heart of our approach to few-shot human body motion retargeting is an implicit volumetric representation of the shape and appearance of the subject depicted in the input images.
In the following sections, we first describe this representation, then outline the network architectures employing it to perform foreground motion transfer and composition into the background environment.
Finally, we discuss the training techniques and loss functions employed to train these networks to perform flexible motion retargeting while still achieving high-fidelity image synthesis results.
The overall architecture is illustrated in Figure~\ref{fig:pipline}.

\subsection{Implicit Volumetric Representation}
Given an input image, our encoder extracts a feature volume or ``bottleneck'' $f \in \mathbb{R}^{N\times D\times H\times W}$, where $D$, $H$, and $W$ are the depth, height, and width of the encoded volumetric grid used to represent the image content.
Each cell in the grid contains an $N$-dimensional feature vector describing the structure and appearance of the local image content corresponding to that region of the volume.
The depth dimension of this volume is aligned with the view direction of the camera, while the height and width correspond to those of the input image.
The feature volume may be passed directly to the image decoder to synthesize an image corresponding to the input (in which case it acts as an auto-encoder), or it may be spatially manipulated in a manner corresponding to the desired transformation of the image content.
Such manipulations include rigid transformations corresponding to camera viewpoint changes, or non-rigid manipulations corresponding to the subject's body motion.

Given the encoded bottleneck $f$ and a dense \emph{flow field} $\mathrm{T}_{s\rightarrow t} \in \mathbb{R}^{3\times D\times H\times W}$ encoding a 3D coordinate per cell in the transformed bottleneck $f'$ pointing to the original volume $f$ that corresponds to the mapping from pose $s$ to $t$, we employ a sampling layer ${\sampler: f, \mathrm{T}_{s\rightarrow t} \rightarrow f'}$ to perform trilinear sampling to produce $f'$.
This flexible sampling mechanism enables a wide variety of spatial manipulations, from rigid transformations for novel view synthesis (\eg simulating camera pose change by rotating the bottleneck) to more complex non-rigid changes (\eg raising the arms of the subject while keeping the rest of the body stationary).

While the flow field $\mathrm{T}_{s\rightarrow t}$ for rigid camera viewpoint changes can be easily computed given the relative camera transformation between $s$ and $t$, non-rigid body pose transformations can be much more complicated. It requires the flow field to have the appropriate sampling location in $f$ for each cell in $f'$ that semantically corresponds to the desired 3D motion of the content extracted from the input image.

Our training process, as described below, in which both rigid viewpoint and non-rigid pose transformations are used, achieves the \emph{spatial disentanglement} required to infer the appropriate flow field and employ it to guide the desired spatial transformation of the image content.

\subsection{Flow-Guided Volumetric Resampling}

\noindent\textbf{Network Architecture Overview.} 
Our human body synthesis branch $\mathrm{G}_\mathrm{fg}$ has three major components: the encoder, which consists of a 2D encoding network $\mathrm{Enc}_\mathrm{2D}$ and a 3D encoding network $\mathrm{Enc}_\mathrm{3D}$; the decoder, which consists of a 3D decoding network $\mathrm{Dec}_\mathrm{3D}$ and a 2D decoding network $\mathrm{Dec}_\mathrm{2D}$; and the flow-guided transformable bottleneck network $\mathrm{F}$.
A source reference image $\mathbf{x}_\mathrm{fg}$, containing only the foreground region from the original image $\mathbf{x}$ (obtained using a pre-trained segmentation model~\cite{chen2018encoder}), is passed through the encoder to obtain the 3D feature representation $f$ as $f$ = $\mathrm{Enc}_\mathrm{3D}(\mathrm{Enc}_\mathrm{2D} (\mathbf{x}_\mathrm{fg}))$.
To warp the features, we estimate the flow using the flow network $\mathrm{F}$ with inputs of $\mathbf{x}_\mathrm{fg}$ and the source/target poses $\mathbf{p}_s, \mathbf{p}_t$, which are obtained with the off-the-shelf pose detection network~\cite{cao2018openpose,alp2018densepose}.
For simplicity, we define this operation as $\mathrm {F}_{s\rightarrow t}(f) \coloneqq \sampler(f, \mathrm{F}(\mathbf{x}_\mathrm{fg}, \mathbf{p}_s,\mathbf{p}_t))$, and generate the synthesized foreground frame using the flow field from $\mathrm{F}$ as:
\begin{equation}\label{eq:synthesis}
\hat{\mathbf{y}}_\mathrm{fg} = \mathrm{Dec}_\mathrm{2D}(\mathrm{Dec}_\mathrm{3D}(\mathrm{F}_{s\rightarrow t}(f))).
\end{equation}

\noindent\textbf{Multi-Resolution Bottleneck.}
To increase the fine-scale fidelity of the synthesized images while retaining the global structure of the target subject, we adopt a multi-resolution representation, with implicit volumetric representations at multiple resolutions, as seen in Figure~\ref{fig:pipline}. We employ skip connections between each 3D encoder and 3D decoder. Therefore, the encoder produces 3D features such that $f=\left\{f_1,f_2,\ldots,f_m\right\}$, where $m$ is the number of resolutions. The skip connections use 3D warping, given the estimated flow, to map content to from the correct region in the encoded bottleneck to the one to be decoded, rather than the direct connections used in prior art~\cite{he2016deep}. Thus, for each feature $f_i$ obtained from the 3D encoder, the 3D decoder receives as input $\mathrm{F}_{s\rightarrow t}(f_i)$.

\noindent\textbf{Multi-View Aggregation.}
Given that a single input image only contains partial information for the depicted human body, we allow for the aggregation of information, represented with our 3D features, from multiple views to improve the synthesized image quality. To accelerate the inference, we take the first source image as the canonical body pose (though this may actually be the subject in any natural pose), and aggregate features from all other images to this pose. In this way, we only perform the aggregation once during inference. More formally, for a total of $N$ source images, the aggregated feature $f_a$ is represented as: 
\begin{equation}\label{eqn:aggregate}
   f_a = \frac{1}{N}(f_1 + \sum_{i=2}^{N}\mathrm{F}_{i\rightarrow1}(f_i)),
\end{equation}
where $f_i$ is the 3D representation from the image $i$ and $\mathrm{F}_{i\rightarrow1}$ is the warping flow from the image $i$ to the first image.

\subsection{Background Modeling}
The background environment is modeled through a separate network $\mathrm{G}_\mathrm{bg}$, to which the source images are fed. Specifically, $\mathrm{G}_\mathrm{bg}$ estimates a background image $\hat{\mathbf{y}}_\mathrm{bg}$ and a confidence map $\hat{\mathbf{w}}$, in which high confidence indicates the foreground, \ie the depicted person, while low confidence indicates the background. The synthesized image $\hat{\mathbf{y}}$ is:
\begin{equation}\label{eqn:composition}
   \hat{\mathbf{y}} = \hat{\mathbf{w}} \cdot \hat{\mathbf{y}}_\mathrm{fg} + (1 - \hat{\mathbf{w}}) \cdot \hat{\mathbf{y}}_\mathrm{bg},
\end{equation}
where $\cdot$ denotes component-wise multiplication of the confidence map with the color channels of the synthesized image, and $\hat{\mathbf{y}}_\mathrm{bg}$ indicates the synthesized background image.

\subsection{Training}
\noindent\textbf{Retargeting Supervision.}
We use a conditional discriminator $\mathrm{D}_\mathrm{fg}$ to determine whether the synthesized foreground image is real or fake. The concatenation of the input image, 2D source/target poses, and target image is sent to the discriminator and we apply the following adversarial loss:
\begin{equation}
\begin{split}
 \mathcal{L}_{\mathrm{D}_\mathrm{fg}}=\mathbb{E}_{\mathbf{p}_t, \mathbf{y}_\mathrm{fg}}[\log \mathrm{D}_\mathrm{fg}( \mathbf{p}_t, \mathbf{y}_\mathrm{fg})] + \\
 \mathbb{E}_{\mathbf{x}_\mathrm{fg},\mathbf{p}_s, \mathbf{p}_t}[\log(1-\mathrm{D}_\mathrm{fg}( \mathbf{p}_t, \mathrm{G}_\mathrm{fg}(\mathbf{x}_\mathrm{fg},\mathbf{p}_s, \mathbf{p}_t)))],
\end{split}
\label{eq:d_fg}
\end{equation}
where $\mathbf{y}_\mathrm{fg}$ is the foreground region from the real image.
The generator is trained to minimize this objective, while the discriminator is trained to maximize it.

Similarly, we have a discriminator $\mathrm{D}_\mathrm{bg}$ that works on the full synthesized with foreground and background:
\begin{equation}
\begin{split}
 \mathcal{L}_{\mathrm{D}_\mathrm{bg}}=\mathbb{E}_{\mathbf{p}_t, \mathbf{y}}[\log \mathrm{D}_\mathrm{bg}(\mathbf{p}_t, \mathbf{y})] + \\
 \mathbb{E}_{\mathbf{x},\mathbf{p}_s, \mathbf{p}_t}[\log(1-\mathrm{D}_\mathrm{bg}(\mathbf{p}_t, \mathrm{G}_\mathrm{bg}(\mathbf{x},\mathbf{p}_s, \mathbf{p}_t)))],
\end{split}
\label{eq:d_bg}
\end{equation}
where $\mathbf{y}$ denotes the real image.

We also use a perceptual loss~\cite{johnson2016perceptual} $\mathcal{L}_\mathrm{vgg}$ between the real and generated images to improve fidelity of the generated images for both the foreground and background:
\begin{equation}\label{eqn:vgg}
\mathcal{L}_\mathrm{per} = \mathcal{L}_\mathrm{vgg}( \mathbf{y}_\mathrm{fg}, \hat{\mathbf{y}}_\mathrm{fg} ) + \mathcal{L}_\mathrm{vgg}( \mathbf{y}, \hat{\mathbf{y}}).
\end{equation}

Additionally, we measure the reconstruction quality of each of the $N$ source images using the aggregated bottleneck. The reconstructed image is generated as $\hat{\mathbf{x}} = \mathrm{Dec}_\mathrm{2D}(\mathrm{Dec}_\mathrm{3D}((f)))$, with no flow required for this auto-encoding. The reconstruction loss $\mathcal{L}_\mathrm{recon}$ is:
\begin{equation}\label{eqn:recon}
\mathcal{L}_\mathrm{recon} =  \frac{1}{N}(\sum_{i=1}^{N}\left \| \hat{\mathbf{x}} - \mathbf{x} \right \|_1).
\end{equation}
  
\noindent\textbf{Mask Supervision.}
We also leverage mask supervision, using the foreground masks, to better supervise the implicit 3D representation modeling. Similar to previous work~\cite{olszewski2019transformable}, we introduce an occupancy decoder to obtain the estimated mask directly from the 3D features. Considering that we have multi-resolution bottlenecks in our architecture, we apply multiple occupancy decoders to get the mask from each of the bottlenecks. Specifically, given the occupancy decoder as $\mathrm{Dec}_\mathrm{occ}$, the estimated mask $\hat{\mathbf{w}}_i$ for the $i$-{th} 3D representation is given as $\hat{\mathbf{w}}_i = \mathrm{Dec}_\mathrm{occ}(f_i)$. The mask loss $\mathcal{L}_\mathrm{mask}$ is thus defined as follows:
\begin{equation}\label{eqn:segment}
  \mathcal{L}_\mathrm{mask} = \frac{1}{N}(\sum_{j=1}^{N} \frac{1}{m}(\sum_{i=1}^{m} \left \| \hat{\mathbf{w}}_{i}^j - \mathbf{w}_i^j \right \|_1)),
\end{equation}
where $\mathbf{w}^j$ is the mask obtained from $j$-{th} source reference image using the pre-trained segmentation network.
 
\noindent\textbf{Unsupervised Random Rotation Supervision.}
We further introduce an unsupervised training technique to help the network learn implicit volumetric representations for the encoded subject with appropriate spatial structure.
Specifically, we apply a random rotation around the vertical axis of the volume to the encoded bottleneck and enforce the corresponding synthesized image to be indistinguishable from the ground-truth views.
The magnitude of this rotation is sampled from a uniform distribution $r\sim U(-180^{\circ}, 180^{\circ})$.
The synthesized image is thus $\hat{\mathbf{y}}_{r} = \mathrm{Dec}_\mathrm{2D}(\mathrm{Dec}_\mathrm{3D}(\mathrm{F}_{s\rightarrow r}(f)))$, which should contain a novel view of the subject performing the same pose as in the source image, where $\mathrm{F}_{s\rightarrow r}$ is the flow field defined by the rigid transformation between the source pose $s$ and the random pose $r$.
However, since there is no ground-truth image corresponding to each random rigid transformation, we introduce a discriminator $\mathrm{D}_\mathrm{rot}$ to match the distribution between real images and $\hat{\mathbf{y}}_{r}$ in an unsupervised manner. We provide the discriminator with the concatenation of the generated image and the original source to better maintain the source identity. We employ an adversarial loss to enforce the rotation constraint on the foreground region as follows:
\begin{equation}
\begin{split}
 \mathcal{L}_{\mathrm{D}_\mathrm{rot}}=\mathbb{E}_{\mathbf{x}_\mathrm{fg},  \mathbf{y}_\mathrm{fg}}[\log \mathrm{D}_\mathrm{rot}(\mathbf{x}_\mathrm{fg},   \mathbf{y}_\mathrm{fg})] + \\
 \mathbb{E}_{\mathbf{x}_\mathrm{fg}, \mathbf{p}_s, r}[\log(1-\mathrm{D}_\mathrm{rot}(\mathbf{x}_\mathrm{fg}, \mathrm{G}_\mathrm{fg}(\mathbf{x}_\mathrm{fg}, \mathbf{p}_s, r)))].
\end{split}
\label{eqn:rotate}
\end{equation} 

\noindent\textbf{Full Objective.}
The full training objective for the entire motion retargeting network is thus given as follows:
\begin{equation}\label{eqn:full}
  \begin{split}
    \mathcal{L}_\mathrm{T} = \lambda_\mathrm{fg} \mathcal{L}_{\mathrm{D}_\mathrm{fg}} + \lambda_\mathrm{bg} \mathcal{L}_{\mathrm{D}_\mathrm{bg}} + \lambda_\mathrm{per} \mathcal{L}_\mathrm{per} + \\
    \lambda_\mathrm{recon} \mathcal{L}_\mathrm{recon} + \lambda_\mathrm{mask} \mathcal{L}_\mathrm{mask} + \lambda_\mathrm{rot} \mathcal{L}_{\mathrm{D}_\mathrm{rot}},
  \end{split}
\end{equation}
where $\lambda_\mathrm{fg}$, $\lambda_\mathrm{bg}$, $\lambda_\mathrm{per}$, $\lambda_\mathrm{recon}$, $\lambda_\mathrm{mask}$, and $\lambda_\mathrm{rot}$ are hyper-parameters to control the weight of each loss function.

\section{Experiments}

\begin{table}[]
    \centering
    \caption{Quantitative results on the iPER~\cite{liu2019liquid} dataset. We compare our method with existing works using the SSIM and LPIPS. }
    \vspace{.25em}
    \label{tab:iPer-num}
    \begin{tabular}{c|cc}
        \toprule
        Method& SSIM$\uparrow$& LPIPS$\downarrow$\\\midrule
        PG2~\cite{ma2017pose} & 0.854 & 0.135  \\
        SHUP~\cite{balakrishnan2018synthesizing} & 0.832 & 0.099  \\
        DSC~\cite{siarohin2018deformable}  & 0.829 & 0.129  \\
        LiquidGAN~\cite{liu2019liquid} &  0.840 & 0.087 \\
        \midrule
        Ours & \textbf{0.868} & \textbf{0.086}  \\ 
        \bottomrule
    \end{tabular}
    \vspace{-1em}
\end{table}
\begin{figure*}[]
\begin{center}
\includegraphics[width=1\linewidth]{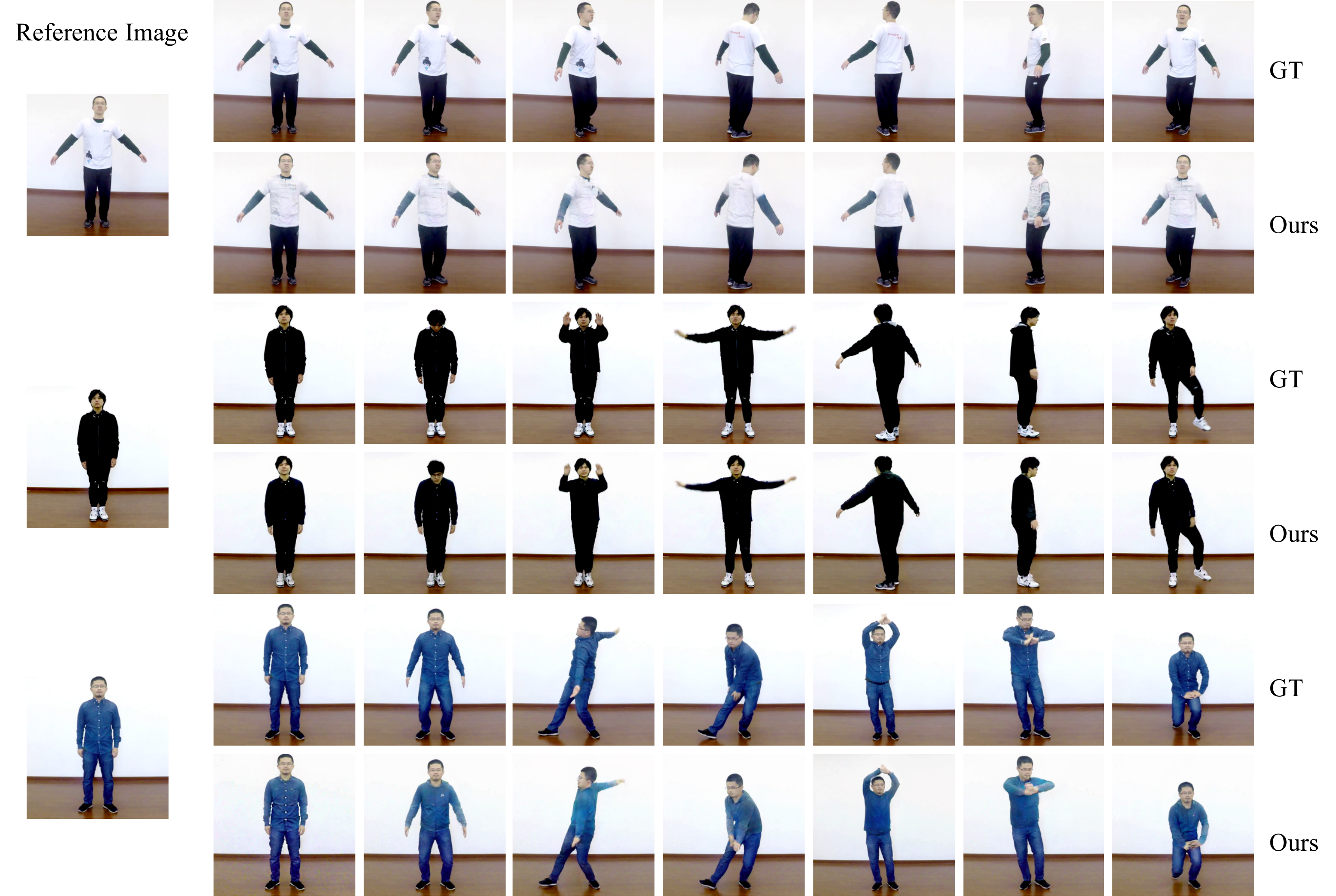}
\caption{Qualitative results on the iPER dataset. The leftmost columns show the reference images. For each example, we show both ground-truth (GT) driving images (odd rows), and the images generated by our method using the pose from the GT images (even rows). Our method can generate realistic images from diverse motion patterns.}\label{fig:iper}
\end{center}
\vspace{-2em}
\end{figure*}
In this section, we evaluate our method on two different datasets, and show qualitative and quantitative comparisons with recent state-of-the-art efforts on human motion generation.
For additional results, including video sequences, please consult the supplemental material.

\subsection{Experimental Setting}

\noindent\textbf{Datasets.}
We adopt two datasets for experiments:
\begin{itemize}[leftmargin=1.5em]
  \setlength\itemsep{-0.25em}
    \item \textbf{iPER}. The first dataset, known as the Impersonator (iPER) dataset, was recently collected and released by LiquidGAN~\cite{liu2019liquid}, and serves as a benchmark dataset for human motion animation tasks.  There are $206$ videos with a total of $241,564$ frames in the dataset. Each video depicts one person performing different actions. We follow the protocol in previous work, splitting the training and testing set in the ratio of 8:2.
    Compared with other datasets~\cite{liu2016deepfashion,zheng2015scalable,zablotskaia2019dwnet}, iPER includes human subjects with widely varying shape, height and appearance, and performing diverse motions.
    \item \textbf{Youtube-Dancing.} We create this dataset by collecting dancing videos from Youtube. It includes $675$ dancing videos with $4,063,453$ frames in total for training, and $200$ videos with $12,965$ frames for testing. The videos are recorded in unconstrained, in-the-wild environments, and consist of more challenging and diverse motion patterns compared with the iPER dataset.
\end{itemize}

\noindent\textbf{Implementation Details.}
We normalize all images to the range $\left [-1, 1\right ]$ for training, and train the networks to synthesize images with a resolution of $256\times256$. We apply the Adam optimizer~\cite{kingma2014adam} with a mini-batch size of $32$ for training. Following previous work~\cite{wang2018high}, we apply multi-scale discriminators, where each discriminator accepts images with the original resolution, $256\times256$ and images with half this resolution, $128\times128$.
The hyper-parameters in Eqn.~\ref{eqn:full} are set to $1$, except for $\lambda_\mathrm{per}=10$.

\noindent\textbf{Evaluation Metrics.}
We use three widely-adopted metrics to evaluate the image synthesis quality.
{The Structural Similarity (SSIM)}~\cite{wang2004image} index measures the structural similarity between synthesized and real images.
{The Learned Perceptual Similarity 
(LPIPS)}~\cite{zhang2018unreasonable} measures the perceptual similarity between these images.
{The Fr$\mathrm{\acute{e}}$chet Inception Distance (FID)}~\cite{heusel2017gans} calculates the distance between two distributions of real and synthesized images, and is commonly used to quantify the fidelity of the synthesized images.

\begin{figure*}[]
\begin{center}
\includegraphics[width=0.95\linewidth]{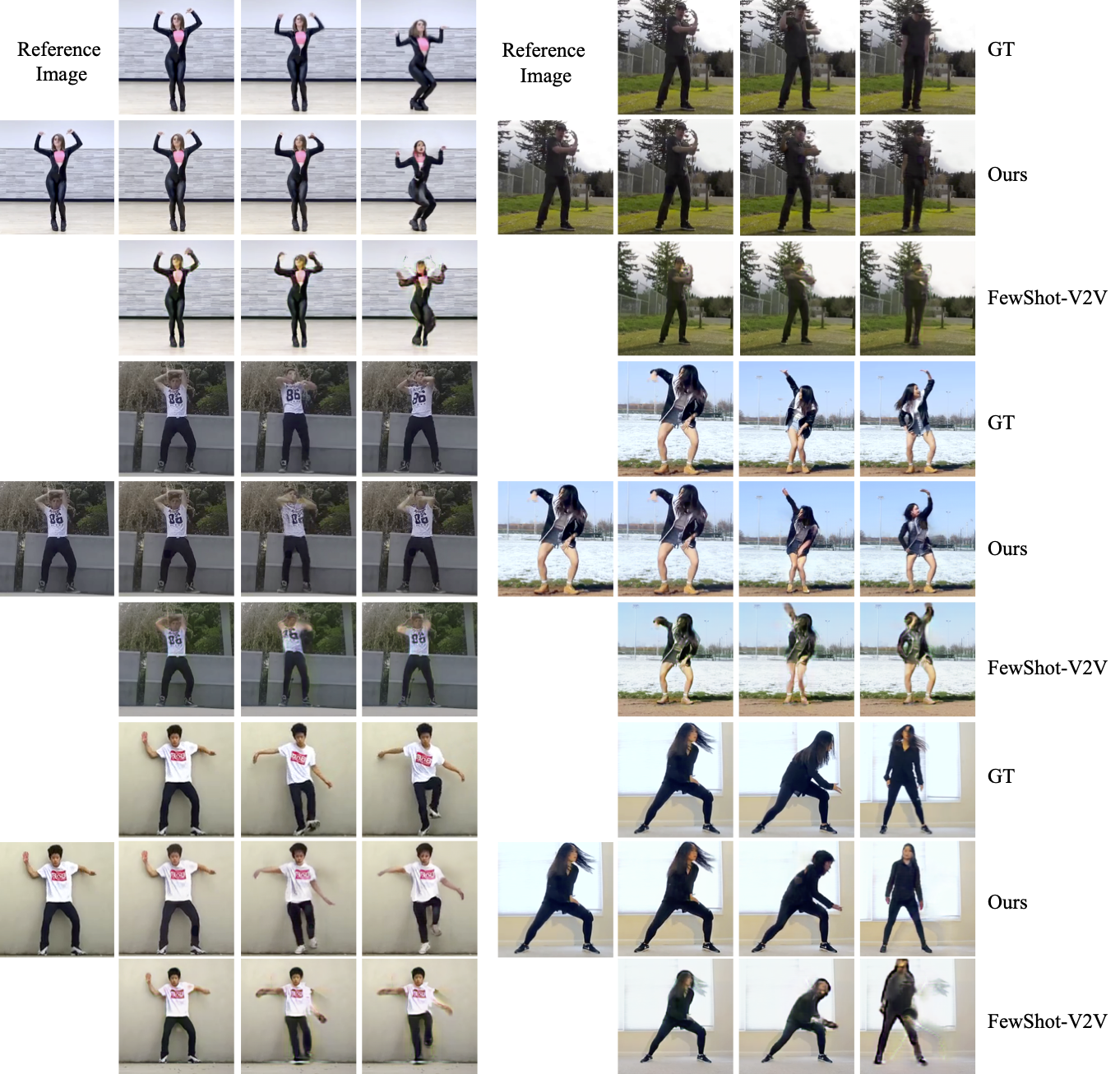}
\caption{Qualitative results on the Youtube-Dancing dataset. The first and fifth column show one reference image from a target person. For each target person, we show short sequences, including the ground-truth driving images, and the images generated by our and FewShot-V2V~\cite{wang2019fewshotvid2vid} using the pose information from the driving images. As seen here, our approach clearly generates more realistic results.}\label{fig:youtube}
\end{center}
\vspace{-2.0em}
\end{figure*}

\begin{table}[]
    \centering
    \caption{Quantitative results on the Youtube-Dancing dataset. We report the SSIM, LPIPS, and FID for both methods.}
    \vspace{.25em}
    \label{tab:youtube}
    \small
    \begin{tabular}{c|ccc}
        \toprule
        Method & SSIM$\uparrow$&LPIPS$\downarrow$&FID$\downarrow$\\\midrule
        FewShot-V2V~\cite{wang2019fewshotvid2vid}  &  0.770 & 0.144  & 48.39 \\
        Ours & \textbf{0.787} & \textbf{0.131} &  \textbf{18.82}\\
    \bottomrule
    \end{tabular}
    \vspace{-.25em}
\end{table}

\subsection{Comparisons and Results}
We first show the experimental results on the iPER dataset. 
Following previous work~\cite{liu2019liquid}, we use three reference images  with different degrees of occlusion from each video to synthesize other images.
We report both the SSIM and LPIPS for our work and existing studies, including PG2~\cite{ma2017pose}, SHUP~\cite{balakrishnan2018synthesizing}, DSC~\cite{siarohin2018deformable}, and LiquidGAN~\cite{liu2019liquid} in Table~\ref{tab:iPer-num}. As shown in the table, we achieve better results using both metrics than state-of-the-art works, indicating higher synthesized image quality with our method. Additionally, we provide qualitative results in Figure~\ref{fig:iper}. We show an example source reference image from a target subject, and the generated sequence using the driving pose information from the ground-truth images. As depicted, our method can generate realistic images with diverse motion patterns.

\begin{table}[]
    \centering
    \caption{Ablation analysis of our proposed architecture on the iPER dataset. Our Full method achieves results that are superior to all other variants.}
    \vspace{.25em}
    \label{tab:ablation}
    \begin{tabular}{c|cc}
        \toprule
        Method & SSIM$\uparrow$ & LPIPS$\downarrow$ \\ \midrule
        w/o Multi-Resolution TBN  & 0.870 &  0.051 \\
        w/o Skip-Connections  & 0.856 & 0.059 \\
        w/o Random Rotations & 0.876 & 0.047 \\
        Full & \textbf{0.878} & \textbf{0.045}\\
        \bottomrule
    \end{tabular}
    \vspace{-.5em}
\end{table}

\begin{table}[]
    \centering
    \caption{Analysis of the use of different numbers of reference images from the target subject for image generation on the iPER dataset. The image quality improves with our method as more reference images are used.}
    \vspace{.25em}
    \label{tab:number_of_img}
    \begin{tabular}{c|cc}
        \toprule
        Method & SSIM$\uparrow$ & LPIPS$\downarrow$ \\ \midrule
         One reference image& 0.878 & 0.045\\
         Two reference images&  0.879 & 0.045\\
        Three reference images&  0.881 & 0.044\\
        Four reference images&  \textbf{0.882} & \textbf{0.043} \\\bottomrule
    \end{tabular}
    \vspace{-1.5em}
\end{table}

We then perform these experiments on our collected Youtube-Dancing dataset. We compare our method with the most recent work, FewShot-V2V~\cite{wang2019fewshotvid2vid}, which can generate motion retargeting videos using one or a few images. For evaluation, we use the first frame from each testing video as a source reference frame and using poses from the other frames to generate images. Besides the SSIM and LPIPS, we also follow FewShot-V2V~\cite{wang2019fewshotvid2vid} and report the FID scores. The results are summarized in Table~\ref{tab:youtube}. As can be seen, our method achieves better results than FewShot-V2V~\cite{wang2019fewshotvid2vid} on each of the three metrics. We also provide sample qualitative results in Figure~\ref{fig:youtube}. We show the source reference image from a target subject, and the generated short sequence from our method and FewShot-V2V using the pose from the ground-truth images. Compared with FewShot-V2V, our method generates more realistic images with fewer artifacts, and the identity and the texture from the target subject are much better preserved.

\subsection{Ablation Study}

\noindent\textbf{Architecture and Training Technique Analysis.}
We conduct an ablation analysis of our network architectures and training strategies. For this experiment, we analyze the effects of each component used to generate the human body, \ie the foreground region. We thus remove the background using an image segmentation model~\cite{chen2018encoder}, and only compute evaluation metrics on the generated foreground region. We use one source reference image from a target subject and preform experiments with the following  setting on iPER:
\begin{itemize}[leftmargin=1.5em]
  \setlength\itemsep{-0.25em}
    \item \emph{w/o Multi-Resolution TBN.} Instead of using multi-resolution bottlenecks, 
    we use a single resolution TBN to encode the implicit 3D representation.
    \item \emph{w/o Skip-Connections.}
    The skip-connections (implemented via flow warping) between the encoded and decoded 3D bottlenecks are removed.
    \item \emph{w/o Random Rotations.} We remove the unsupervised random rotation supervision applied to the encoded TBN and its associated loss function, defined in Eqn.~\ref{eqn:rotate}.
\end{itemize}
The quantitative results, including the SSIM and LPIPS,  are shown in Table~\ref{tab:ablation}. We can see that each component is beneficial and that our full method (\emph{Full}) achieves the best results.

\noindent\textbf{Multi-View Aggregation Analysis.}
As our method can leverage multiple reference images from the target to perform multi-view aggregation, we conduct experiments to analyze the effect of the number of reference images on the synthesis result using the iPER dataset. The experimental results, presented in Table~\ref{tab:number_of_img}, demonstrate that using more reference images improves the quality of the synthesized human bodies.

\section{Conclusion}
Our approach to few-shot human motion retargeting exploits advantages of 3D representations of human body while avoiding limitations of the more straightforward prior methods.
Our implicit 3D representation, learned via spatial disentanglement during training, avoids pitfalls of standard geometric representations such as dense pose estimations or template meshes, which are limited in their expressive capacity and for which it is impossible to obtain accurate ground-truth in unconstrained conditions. However, it allows for 3D-aware motion inference and image content manipulation, and attains state-of-the-art results on challenging motion retargeting benchmarks.
Though we require 2D human poses, our approach could be extended to allow for more general motion retargeting for images of articulated animals given their 2D poses.


{\small
\bibliographystyle{ieee_fullname}
\bibliography{main}
}

\end{document}